\documentclass[sigconf,nonacm]{acmart}

\usepackage{algorithm}
\usepackage{url}
\usepackage{booktabs}
\usepackage{algpseudocode}
\usepackage{graphicx}
\usepackage{textcomp}
\usepackage{subfigure}
\usepackage{multirow}
\usepackage{xcolor}

\AtBeginDocument{%
  }

\setcopyright{acmlicensed}
\copyrightyear{2018}
\acmYear{2018}
\acmDOI{XXXXXXX.XXXXXXX}
\acmConference[Conference acronym 'XX]{Make sure to enter the correct
  conference title from your rights confirmation email}{June 03--05,
  2018}{Woodstock, NY}
\acmISBN{978-1-4503-XXXX-X/2018/06}

\author{Zhenyu Liu}
\authornote{Both authors contributed equally to this work.}
\affiliation{%
\institution{Rensselaer Polytechnic Institute}
\city{Troy}
\state{NY}
\country{USA}
}

\author{Yunzhen Liu}
\authornotemark[1]
\affiliation{%
\institution{University of Massachusetts Amherst}
\city{Amherst}
\state{MA}
\country{USA}
}

\author{Zehao Fan}
\affiliation{%
\institution{Rensselaer Polytechnic Institute}
\city{Troy}
\state{NY}
\country{USA}
}

\author{Garrett Gagnon}
\affiliation{%
\institution{Rensselaer Polytechnic Institute}
\city{Troy}
\state{NY}
\country{USA}
}

\author{Yayue Hou}
\affiliation{%
\institution{Rensselaer Polytechnic Institute}
\city{Troy}
\state{NY}
\country{USA}
}

\author{Nan Wu}
\affiliation{%
\institution{George Washington University}
\city{Washington}
\state{DC}
\country{USA}
}

\author{Yangwook Kang}
\affiliation{%
\institution{Samsung Semiconductor}
\city{San Jose}
\state{CA}
\country{USA}
}

\author{Liu Liu}
\affiliation{%
\institution{Rensselaer Polytechnic Institute}
\city{Troy}
\state{NY}
\country{USA}
}

\begin{document}

\title{Bandwidth-Efficient Adaptive Mixture-of-Experts via Low-Rank Compensation}

\begin{abstract}



Mixture‑of‑Experts (MoE) models scale capacity via sparse activation but stress memory and bandwidth. Offloading alleviates GPU memory by fetching experts on demand, yet token‑level routing causes irregular transfers that make inference I/O‑bound. Static uniform quantization reduces traffic but degrades accuracy under aggressive compression by ignoring expert heterogeneity. We present Bandwidth‑Efficient Adaptive Mixture‑of‑Experts via Low‑Rank Compensation, which performs router‑guided precision restoration using precomputed low‑rank compensators. At inference time, our method transfers compact low‑rank factors with Top‑$n$ ($n<k$) experts per token and applies compensation to them, keeping others low‑bit. Integrated with offloading on GPU and GPU–NDP systems, our method delivers a superior bandwidth–accuracy trade‑off and improved throughput.
\end{abstract}



\maketitle

\section{Introduction}


Mixture-of-Experts (MoE) has become a widely adopted architecture for scaling model capacity without a proportional increase in computation \cite{jiang2024mixtral, liu2024deepseek,yang2025qwen3,  shazeer2017outrageously,muennighoff2024olmoe}. By activating only a small subset of experts per token, MoEs achieve efficient conditional computation and have been deployed in large language models (LLMs) such as Mixtral 8x7B \cite{jiang2024mixtral} and DeepSeek-MoE \cite{liu2024deepseek}. However, this scalability introduces substantial memory pressure. For instance, Mixtral 8x7B \cite{jiang2024mixtral} requires 96.8 GB of loading memory and reaches a peak memory usage of 112.6 GB during inference, which already exceeds the 80 GB capacity of an NVIDIA A100 GPU. Such high memory demand has become a major obstacle that hinders the practical deployment of large MoE models. Consequently, practical deployment typically relies on offloading—keeping inactive experts in secondary memory and fetching active ones on demand \cite{eliseev2023fast}.

While offloading enables large-model inference under tight device memory, the token-level routing that defines MoE introduces irregular and latency-dominant parameter movement across the PCIe (or NVLink) boundary. The GPU frequently stalls waiting for expert weights, pushing execution into an I/O-bound regime where end-to-end latency is governed by bandwidth rather than compute.

To relieve this bandwidth bottleneck, quantization compresses offloaded experts to lower bit-widths, reducing transfer volume and moving execution toward the compute-bound region \cite{eliseev2023fast}. However, static, uniform quantization treats all activated experts identically and fails to capture the token-level dynamics of expert importance (Figure \ref{fig:decoding_pattern}). Router scores are highly skewed across tokens (e.g., the Top‑1 expert dominates in Mixtral models; See Figure~\ref{fig:router_score}), so applying the same aggressive precision to every expert disproportionately harms the dominant expert and degrades accuracy at high compression ratios.


\begin{figure}[t]
  \centering
\includegraphics[width=0.48\textwidth]{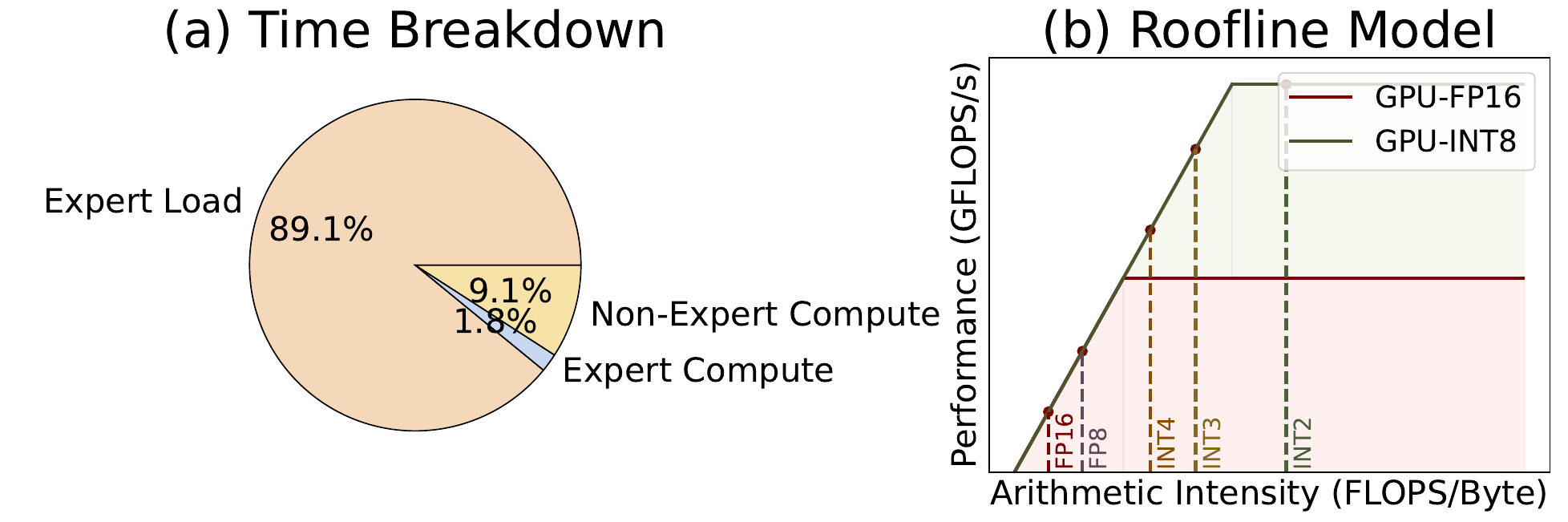} 
  \caption{MoE inference time breakdown and roofline.}
  \label{fig:data-breakdown} 
\end{figure}

We address this mismatch with a runtime, router-guided error compensation mechanism that preserves bandwidth savings while restoring fidelity only where it matters. Specifically, we precompute per-expert low-rank compensators offline. During inference, the router identifies the Top‑$n$ expert(s) per token; for these selected experts, we transfer compact low-rank factors and reconstruct a corrected weight by adding a low-rank update to the quantized parameters on-the-fly. Non‑selected experts remain in low precision and incur no compensation overhead. This design aligns bandwidth expenditure with token-level importance and mitigates quantization error without reverting to full-precision transfer.

In this paper, we present our approach, Bandwidth‑Efficient Adaptive Mixture‑of‑Experts via Low‑Rank Adaptive Compensation. Our method couples offloading with token-aware precision restoration to reduce data movement while maintaining accuracy. Our contributions are threefold:  
\begin{itemize}

\item \textbf{Offline Low-Rank Compensation for Efficient Restoration.}  
We introduce different low-rank compensation modules for experts that enable accurate on-the-fly reconstruction of high-precision experts under bandwidth budgets.

\item \textbf{Token-Level Router-Guided Precision Adaptation.}
We leverage router scores to dynamically restore precision per token, high precision only for the dominant Top-$n$ expert while keeping others quantized.

\item \textbf{System-Level Bandwidth–Accuracy Implementation.}  
We explore our method in both GPU-only and GPU--NDP (Near Data Processing) systems, systematically analyzing the throughput–accuracy trade-off and demonstrating substantial gains in inference efficiency.
\end{itemize}

\section{Background and Motivation}

In this section, we provide the background on MoE inference and the motivation of our method. Firstly, we introduce the MoE architecture and shows the memory-bound of inference under offloading scenario in Section \ref{sec:moe}. Then, we discuss the dynamics of MoE inference in Section \ref{sec:dynamics}. Finally, in Section \ref{sec:compensator}, we motivate our compensator design.

\subsection{Mixture-of-Experts Architecture}
\label{sec:moe}

The MoE architecture scales model capacity by introducing multiple feed-forward experts while activating only a small subset for each token. A gating network computes routing logits through a function $G(\mathbf{x})$, followed by softmax, and selects the top-$k$ experts with the highest routing scores. Formally, for an input token representation $\mathbf{x}$, the normalized routing weight for expert $i$ is
$w_i = \mathrm{softmax}(G(\mathbf{x}))$
and the MoE layer output is
$
    y = \sum_{i \in \mathrm{TopK}(w, k)} w_i \, E_i(\mathbf{x}),
$
This conditional computation allows the total number of experts $N$ to scale without proportionally increasing computation, since only $k \ll N$ experts are active at each step. As model capacity grows, however, the aggregate expert parameters exceed GPU memory capacity, making offloading a necessity during deployment.

To quantify the bottlenecks of offloaded MoE inference, we profile Mixtral-Offloading~\cite{eliseev2023fast} on an NVIDIA H100 PCIe GPU with an AMD EPYC 9334 CPU. We extend the framework to support \texttt{FP16} expert offloading, where expert weights reside in CPU memory and are fetched to the GPU on demand. As shown in Figure~\ref{fig:data-breakdown} (a), the majority of inference time is consumed by host–device data movement, whereas expert computation accounts for only a small portion of the runtime, confirming that offloaded MoE inference is fundamentally memory-bound.

A natural approach for reducing transfer cost is to compress expert parameters using quantization. Figure~\ref{fig:data-breakdown} (b) illustrates that lowering expert precision from 16-bit to 3-bit and 2-bit proportionally reduces parameter size and shifts the operation intensity upward, pushing inference away from the memory-bound. However, unlike dense models, MoE is highly sensitive to low-bit quantization, and the resulting approximation error can substantially degrade accuracy \cite{huang2024mixture, huang2025milo,chitty2025mopeq}. This raises a key question: \textit{how can we retain the bandwidth benefits of low-bit quantization while effectively mitigating the approximation error it introduces?}

\subsection{Dynamics of MoE Inference}
\label{sec:dynamics}

\begin{figure}[t]
  \centering
\includegraphics[width=0.5\textwidth]{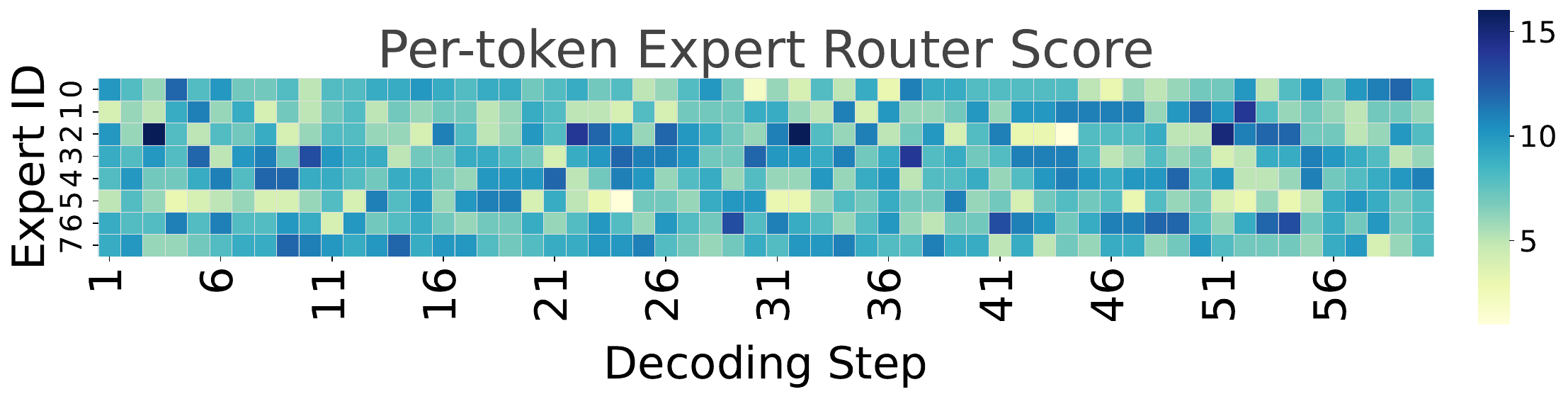}
  \caption{Decoding expert router patterns.}
  \label{fig:decoding_pattern}
\end{figure}

\begin{figure}[t]
  \centering

  \includegraphics[width=0.5\textwidth]{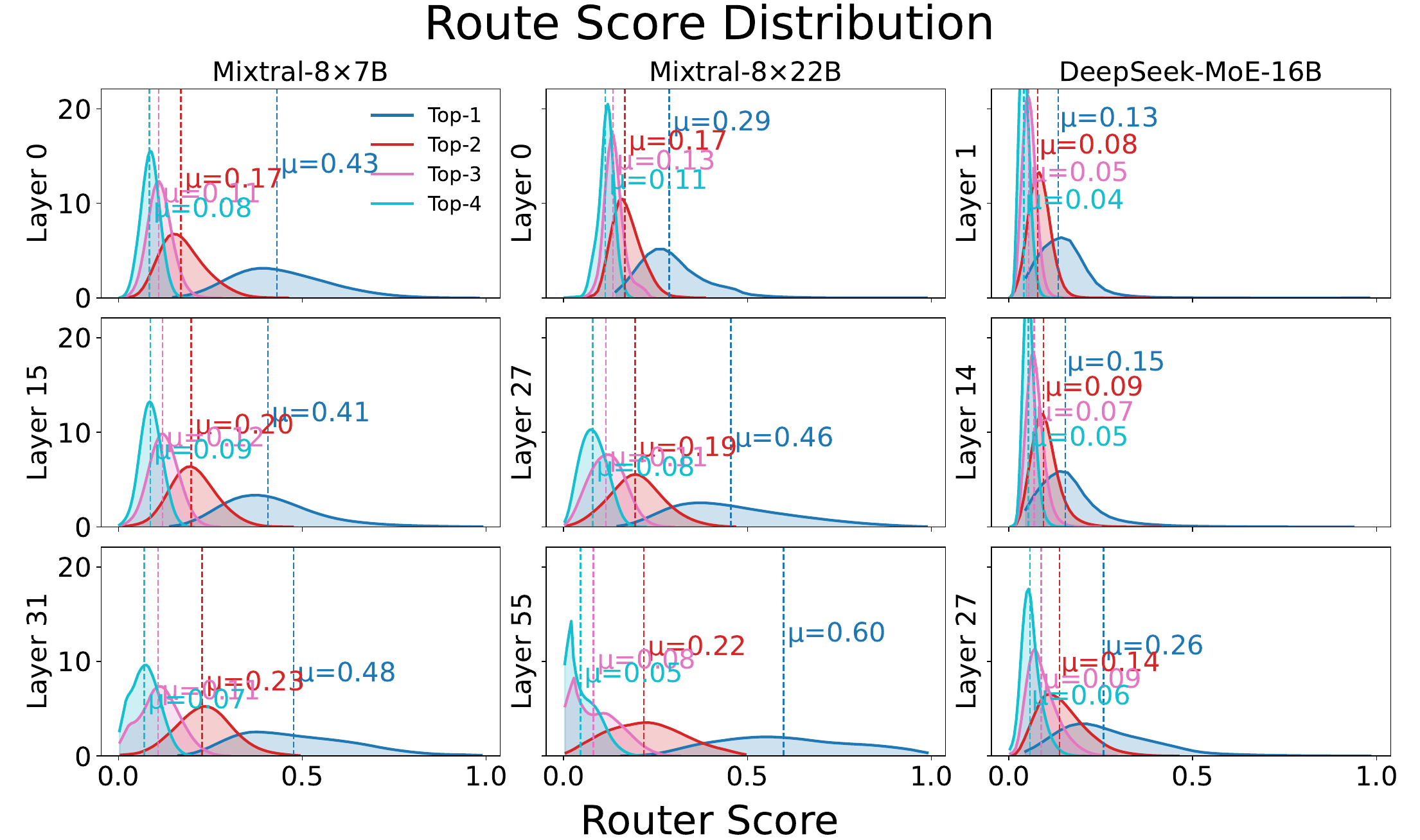}

  \caption{Router score distribution on C4 calibration dataset.}
  \label{fig:router_score}
\end{figure}

To understand why MoE models incur larger quantization error, we take a closer look at their inference behavior. We examine the inference dynamics of Mixtral-8$\times$7B \cite{jiang2024mixtral} and observe that their behavior varies substantially across decoding steps. As shown in Figure~\ref{fig:decoding_pattern}, the set of activated experts changes irregularly during decoding, and the relative importance of each expert shifts over time. This variability makes it challenging to apply uniform quantization, since any expert may become activated at one step.

These dynamics stem from the router, whose fluctuating decisions cause experts to vary in importance across decoding steps. This insight motivates exploiting router information to enable high-quality, importance-aware quantization. We randomly select 80K tokens in the C4 dataset \cite{raffel2020exploring} and get the statistics of routing scores. As shown in Figure~\ref{fig:router_score}, routing scores are strongly skewed toward the top-$n$ experts ($n<k$). For Mixtral-8$\times$7B \cite{huang2024mixture}, the top-1 expert has a mean score of 0.41–0.48, while Top-2 drops sharply to 0.17–0.20. Mixtral-8$\times$22B \cite{huang2024mixture} exhibits an even larger separation, with top-1 reaching 0.46–0.60 compared to 0.17–0.22 for Top-2 and below 0.10 for others, and a similar pattern appears in DeepSeek-MoE-16B\cite{liu2024deepseek}. Across all models, the top-$n$ experts receive higher scores than the remaining activated experts, indicating that quantization errors on them dominate overall quality. This motivates leveraging router information to selectively preserve precision for the top-$n$ experts. But how can we restore the top-$n$ precision efficiently without reintroducing large transfer overhead? This question leads to our low-rank compensation mechanism, which will be introduced next.

\subsection{Low-rank Compensation for Experts}
\label{sec:compensator}

\begin{figure}[!ht]
  \centering
  \includegraphics[width=0.45\textwidth]{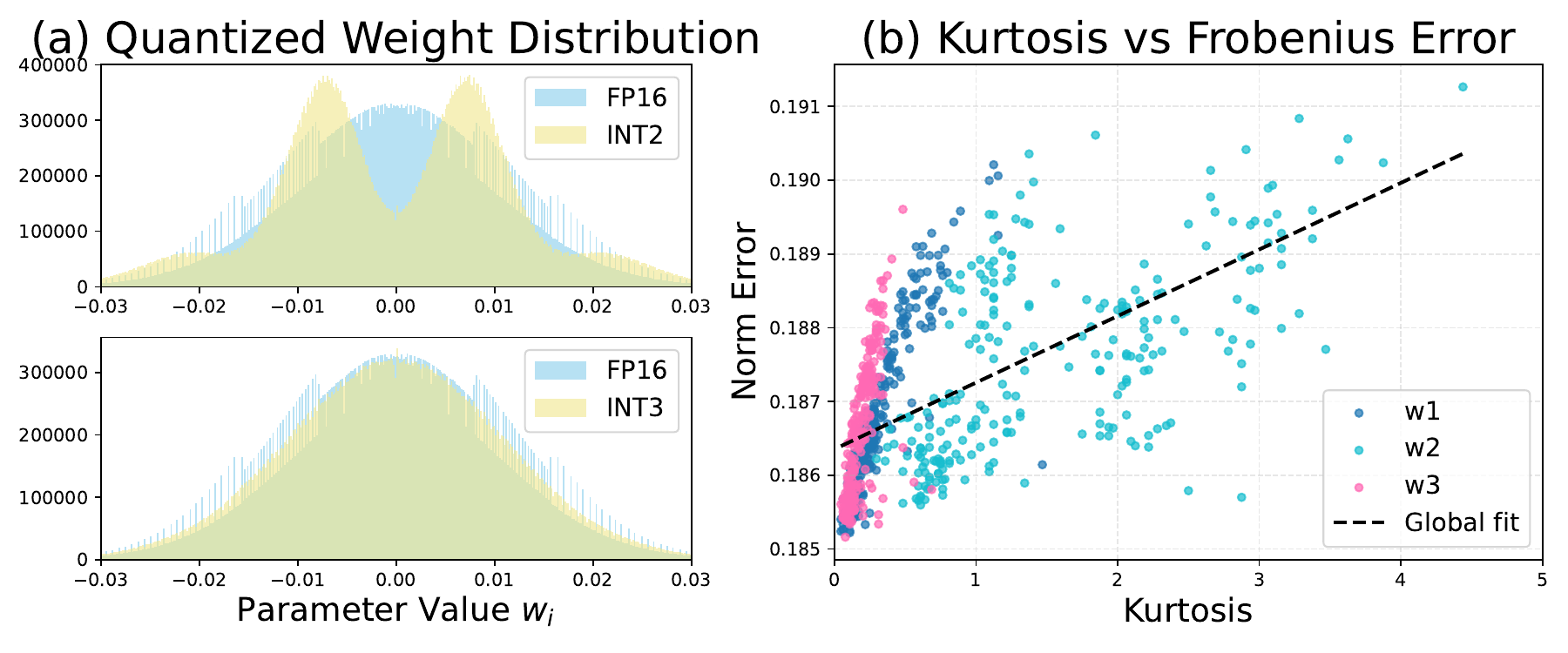}
  \caption{(a) Residual errors introduced by low-bit quantization, which can be effectively restored using a low-rank compensator. (b) Relationship between kurtosis and quantization error, showing that kurtosis is positively related to quantization error, so that higher-kurtosis experts require higher-rank compensation.}
  \label{fig:low_rank}
\end{figure}

Based on the previous analysis, we further investigate how to efficiently restore top-$n$ experts accuracy under low-bit quantization. Prior studies have shown that residual correction, known as \textit{quantize-then-compensate} \cite{yao2024exploring, huang2025milo, zhang2024lqer, saha2024compressing}, can effectively recover the accuracy loss caused by quantization in LLMs. This approach adds a compact low-rank compensator to quantized weights, providing an efficient way to correct approximation errors. Formally, for the weight matrix of each expert ($W_i \in \mathbb{R}^{m \times n}$), this method first apply a quantization operator $Q(\cdot)$ to obtain its low-bit form, and we denote the corresponding dequantized weight by $Q^{-1}(\cdot)$. The quantization residual is then
\[
E_i = W_i - Q^{-1}(Q(W_i)).
\]

To reduce this residual error, this method constructs a lightweight rank-$r_i$ approximation:
$
E_i \approx U_i V_i, 
U_i \in \mathbb{R}^{m \times r_i},\ 
V_i \in \mathbb{R}^{r_i \times n}.
$ 
The naive \textit{quantize-then-compensate} adopts a two-stage procedure:
\begin{enumerate}
    \item \textbf{Quantization.} Quantize all expert weights once to obtain $Q(W_i)$ and $Q^{-1}(Q(W_i))$ for every expert.
    \item \textbf{Uniform SVD per expert.} Manually choose a fixed rank $r$ and, for every expert, perform a truncated SVD on its residual $E_i = W_i - Q^{-1}(Q(W_i))$ to obtain a rank-$r$ factorization $E_i \approx U_i V_i$, which is then stored offline.
\end{enumerate}


However, when naively applied to MoE, the quantization error of expert weights is \emph{not} uniform: 
We observe a correlation between the Frobenius norm $\frac{\|E_i\|_F}{\|W_i\|_F}
=
\frac{\|\,W_i - Q^{-1}(Q(W_i))\,\|_F}{\|W_i\|_F}$  and the kurtosis of the corresponding weight. As shown in Figure \ref{fig:low_rank}, experts with larger kurtosis consistently exhibit larger quantization error, which aligns with previous work \cite{chmiel2020robust,akhondzadeh2025kurtail,huang2025milo}.
This heterogeneity implies that assigning a uniform rank with each expert is a suboptimal, as many experts are over-compensated while those with high kurtosis remain under-compensated. This motivates our rank allocation design in Section \ref{sec:rank_allocation}.

Together, these insights directly motivate our design, enabling us to retain the bandwidth benefits of low-bit quantization while effectively mitigating the approximation error.

\section{Method}
\label{sec:method}

\begin{figure}[h]
  \centering
\includegraphics[width=0.45\textwidth]{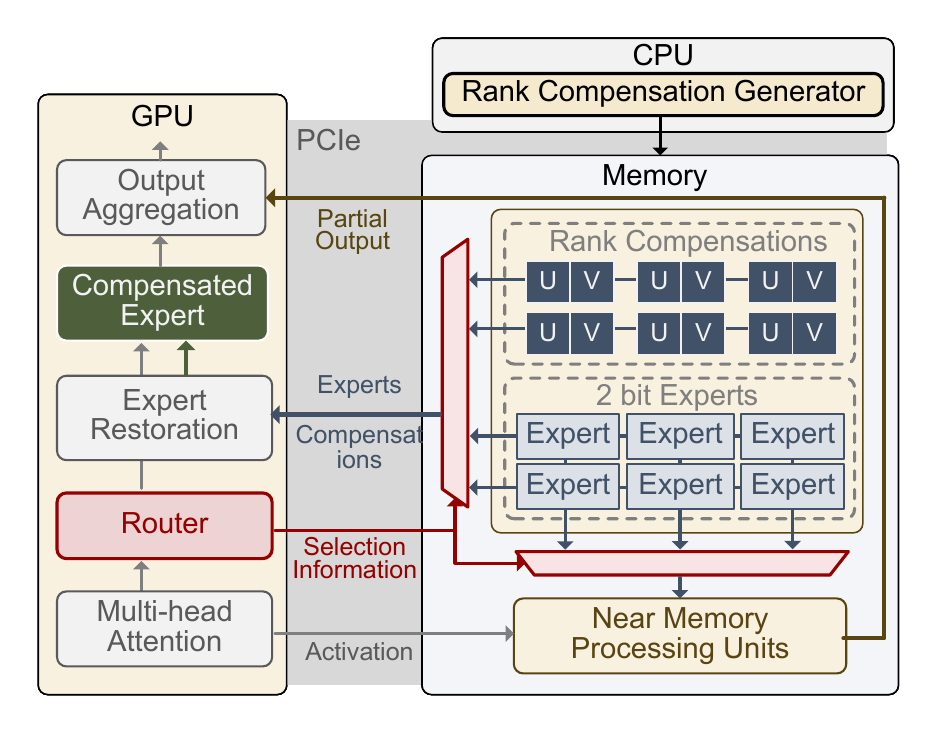}
  \caption{Overview of our method.}
  \label{fig:method_overview}
\end{figure}

In this section, based on our motivation, we first present our kurtosis-guided rank allocation strategy in Section \ref{sec:rank_allocation}. 
Next, we provide the details of the router-guided dynamic precision restoration in Section \ref{sec:beam_dynamic}. As shown in Figure \ref{fig:method_overview}, our method implements a router-guided, selective precision restoration scheme that compensates only the top-$n$ experts to achieve bandwidth-efficient MoE inference.

\subsection{Kurtosis-Guided Rank Allocation}
\label{sec:rank_allocation}

Here, we introduce how better to match the heterogeneous compensation rank of MoE experts, and specifically, we adopt a two-step procedure that leverages this empirical correlation.

\smallskip
\noindent\textbf{Step 1: Offline kurtosis computation and rank assignment.}
For each expert weight matrix $W_i$ (each projection such as $w1/w2/w3$), we compute its kurtosis over all elements:
\[
\kappa_i 
= \frac{1}{d} \sum_{j=1}^{d} 
\frac{(w_{i,j} - \mu_i)^4}{\sigma_i^4},
\]
where $d = m \times n$ is the number of parameters, 
$\mu_i$ and $\sigma_i$ are the mean and standard deviation of entries in $W_i$, 
and $w_{i,j}$ denotes the $j$-th element.
Empirically, experts with larger $\kappa_i$ produce larger quantization residuals $\frac{\|E_i\|_F}{\|W_i\|_F}$, 
indicating that they need more rank for compensation.

We discretize the candidate ranks into a small number of \texttt{bucket},
\[
\texttt{bucket} = \{0,\, 16,\, 32,\, 128,\, 256,\, 512,\, 1024\},
\]
and assign each expert a rank $r_i \in \texttt{bucket}$ using a simple greedy policy.
Let $N$ be the number of experts and $R_{\text{avg}}$ be the target \emph{average} rank budget.
Our goal is to satisfy
$
\frac{1}{N} \sum_{i=1}^{N} r_i = R_{\text{avg}}
$.
To determine $\{r_i\}$, we first sort all experts in descending order of their kurtosis values $\kappa_i$.
We then traverse this sorted list and, for each expert, assign the largest feasible \texttt{bucket} value such that the global constraint$
\sum_{i=1}^{N} r_i \le N\,R_{\text{avg}}$
is not violated.
In effect, high-kurtosis experts receive larger ranks (e.g., $256$--$1024$), 
while low-kurtosis experts are allocated small ranks or even $r_i = 0$.
This kurtosis-guided bucketing aligns rank capacity with the observed quantization difficulty of each expert.

\smallskip
\noindent\textbf{Step 2: HQQ optimization and one-time SVD.}
Given the assigned rank $r_i$ for each expert, our method performs low-bit quantization with HQQ-style weight optimization to obtain the final quantized expert $Q(W_i)$ and its dequantized form $Q^{-1}(Q(W_i))$, and then computes a single truncated SVD on the residual:
\[
U_i, S_i, V_i^\top = \mathrm{SVD}_{r_i}(E_i),\qquad
E_i = W_i - Q^{-1}(Q(W_i)).
\]
We further reparameterize $U_i$ and $V_i$ via
\[
U_i \leftarrow U_i \sqrt{S_i},\qquad 
V_i \leftarrow \sqrt{S_i}\, V_i^\top,
\]
and store the INT3 quantized factors $(\hat U_i, \hat V_i)$ as the low-rank compensator.
All quantized weights and their associated low-rank modules are precomputed offline based on the above procedure and stored for efficient retrieval during MoE inference.




\subsection{Router-Guided Error Compenstation}
\label{sec:beam_dynamic}
 
A straightforward solution for restoring expert precision is to transfer
all compensators $(U_i, V_i)$ of all experts to the GPU. However, based on our insight in Section~\ref{sec:dynamics}, MoE routing
is highly uneven, and only a small subset of experts meaningfully contributes to each token. Therefore, transferring compensators for all experts is unnecessary and incurs substantial bandwidth and memory overhead, particularly as the
number of activated experts grows.

Our method restores precision selectively by leveraging router predictions. Given the sorted router scores for token $t$, we select the top-$n$ experts with the highest routing scores. Only these top-$n$ experts transfer their compensators to the GPU and
reconstruct their compensated weights, while all remaining experts use their lightweight dequantized forms without compensation. This router-guided mechanism ensures that precision is restored only for experts that are truly important for the current token.

During inference, the quantized weights and compensators of the selected
top-$n$ experts are transferred together to the GPU. The compensated weights are reconstructed on-device:
\[
\hat{W}_e = Q^{-1}(Q(W_e)) + U_e V_e,
\]
while all other experts remain in their quantized form $Q^{-1}(Q(W_e))$.
This partial restoration provides token-level adaptivity: compensation
cost is only for experts that matter for the current token, further reducing data movement and computation overhead.



\section{Evaluation}

In this section, we evaluate our method across accuracy, performance, and ablation study of our methods. We begin with the experimental setup in Section~\ref{sec:exp:setup}, covering model architectures, datasets, hardware environments, and baselines. We then present accuracy results in Section~\ref{sec:exp:acc} to demonstrate the effectiveness of our method's strategy on perplexity and reasoning benchmarks. Next, Section~\ref{sec:exp:efficiency} reports end-to-end latency and throughput under both GPU-only and GPU–NDP cases, highlighting the system-level efficiency improvements. Finally, Section~\ref{sec:exp:ablation} provides detailed ablation studies that validate the impact of routing-aware expert selection, rank configuration, and kurtosis-guided rank allocation.

\begin{figure}[t]
  \centering
  \includegraphics[width=0.49\textwidth]{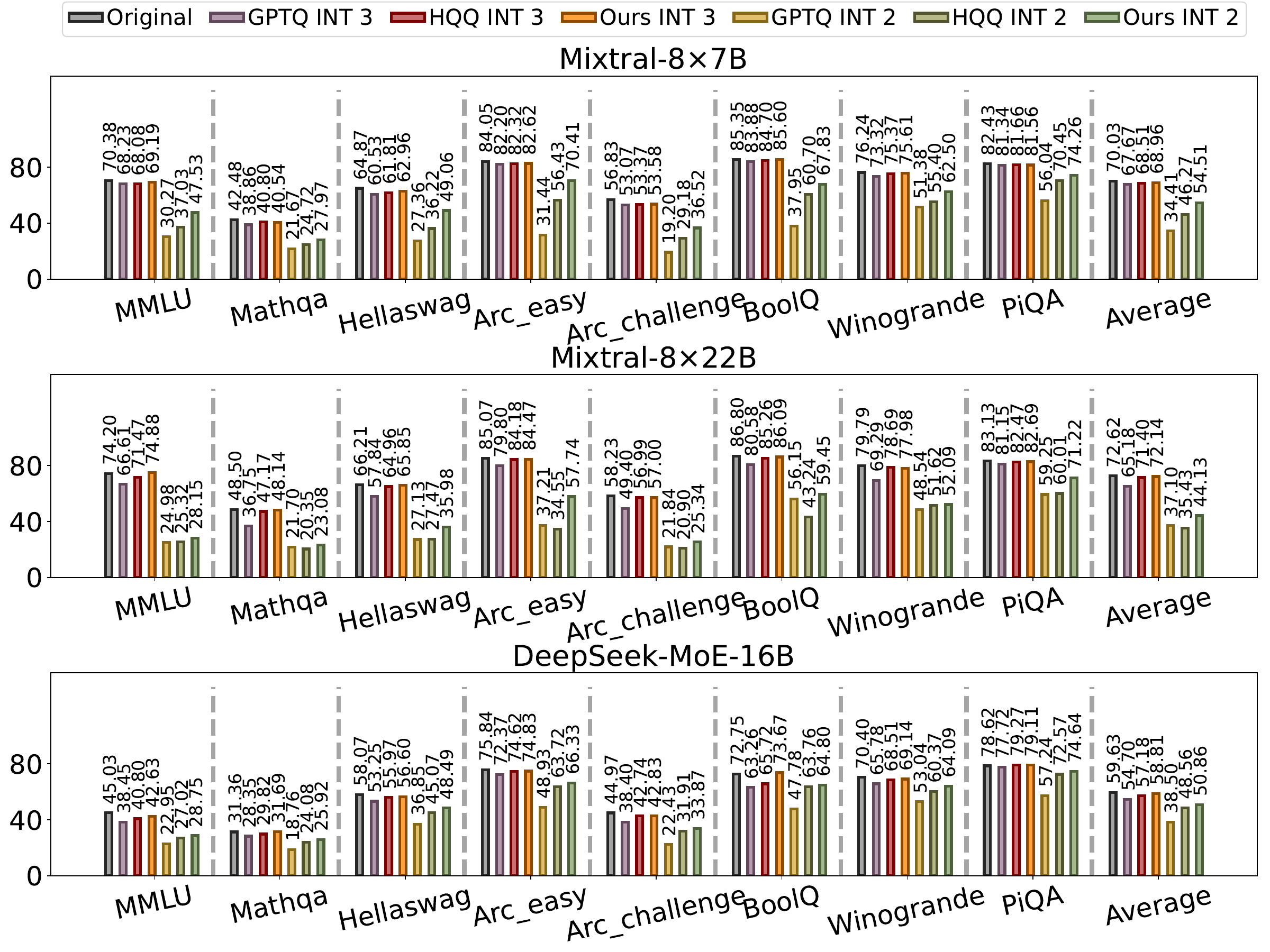}
  \caption{Accuracy results of commonsense reasoning.}
  \label{fig:acc_results}
\end{figure}

\begin{figure*}[ht]
  \centering
  \includegraphics[width=0.98\textwidth]{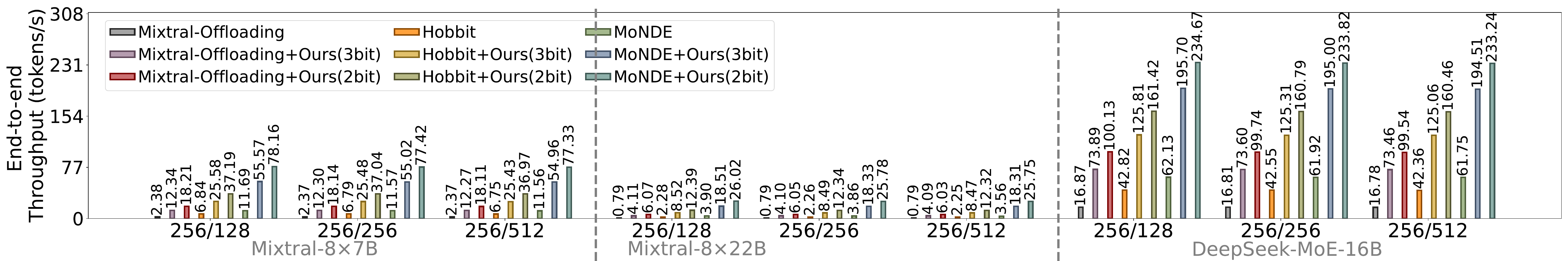}
  \caption{System performance evaluation of GPU-only and GPU-NDP offloading systems.}
  \label{fig:thr_results}
\end{figure*}

\subsection{Experimental Setup}
\label{sec:exp:setup}

\textbf{Models.} We evaluate three representative MoE models. First, we adopt Mixtral-8$\times$7B \cite{jiang2024mixtral}, which activates 2 out of 8 experts for each token. To assess scalability with larger expert dimensions, we further evaluate Mixtral-8$\times$22B \cite{jiang2024mixtral} as a higher-capacity workload. Finally, to study architectures with more experts but smaller per-expert size, we include DeepSeek-MoE-16B \cite{liu2024deepseek}, which contains 64 experts, activating 2 shared experts and 6 independent ones per token.
A detailed comparison of model configurations is summarized in Table~\ref{tab:model-config}.

\textbf{Datasets.}  We evaluate zero-shot commonsense reasoning on seven benchmark tasks (Mathqa \cite{amini2019mathqa}, Hellaswag\cite{zellers2019hellaswag}, Arc-eacy and Arc-challenge\cite{clark2018think}, BoolQ\cite{clark2019boolq}, Winogrande\cite{sakaguchi2021winogrande} as well as PiQA\cite{bisk2020piqa}) and 5-shot performance on MMLU \cite{hendrycks2020measuring}, using the EleutherAI LM Evaluation Harness~\cite{eval-harness}. Additionally, we use the WikiText2 dataset to measure perplexity (PPL) (lower is better) as a proxy for token prediction capability, mainly in the ablation study. These datasets together provide a balanced evaluation of both reasoning ability and generative quality.

\begin{table}[ht]
\centering
\caption{Inference configs of evaluated MoE models}
\label{tab:model-config}
\resizebox{\columnwidth}{!}{%
\begin{tabular}{|l|c|c|c|c|c|c|}
\hline
\textbf{Model} & \textbf{Hidden} & \textbf{Layers} & \textbf{Experts} & \textbf{Top-k} & \textbf{Experts Params.} & \textbf{Params.} \\ \hline
Mixtral-8$\times$7B \cite{jiang2024mixtral}   & (4096, 14336)  & 32 & 8  & 2 & 45.1B  & 46.7B  \\ 
Mixtral-8$\times$22B \cite{jiang2024mixtral}  & (6144, 16384)  & 56 & 8  & 2 & 135.5B & 140.6B \\
DeepSeekMoE-16B$^{\ast}$ \cite{liu2024deepseek} & (2048, 11008) & 28 & 64 & 6 & 15.5B  & 16.4B  \\ \hline
\end{tabular}}
\\[0.25em]
\begin{minipage}{\columnwidth}
\footnotesize
$^{\ast}$DeepSeek-MoE-16B additionally includes two shared experts per MoE layer, which are activated for every token during inference.
\end{minipage}
\end{table}

\textbf{Methodology.} Our experiments are conducted under two deployment scenarios: GPU-only and GPU–NDP hybrid systems, adapted from the same GPU–NDP system configuration used in MoNDE~\cite{kim2024monde} based on Ramulator~\cite{kim2015ramulator}.
In the GPU-only setting, the system consists of an NVIDIA H100 GPU (989.4 TFLOPS, 80GB HBM3) and DDR memory. All experts are loaded to the GPU on demand via PCIe. 
In the GPU–NDP setting, we configure our system as an NVIDIA H100 GPU and a NDP device (512GB/s Bandwidth, 512GB Capacity). Low-bit experts with lightweight computation are directly executed on NDP device, thereby further reducing data movement. We set the input length to 256 and evaluate different output token configurations to measure the end-to-end throughput under both environments.

\textbf{Baselines.} For accuracy evaluation, we compare against state-of-the-art quantization methods, including GPTQ \cite{frantar2022gptq}, and HQQ \cite{badri2023hqq}.
For performance evaluation, we use Mixtral-Offloading \cite{eliseev2023fast} and HOBBIT \cite{tang2024hobbit} as baselines in the GPU-only setup, and MoNDE \cite{kim2024monde} as the baseline for GPU–NDP deployment. These comparisons cover both algorithmic and system-level perspectives.

\subsection{Accuracy Evaluation}
\label{sec:exp:acc}

In our method, we manually configure the compensation settings for each model family. For Mixtral models, we set the average rank budget to 32, and dynamically adjust the precision only for the top-1 activated expert, reflecting the top-1 expert dominates the router score. In contrast, DeepSeek-MoE-16B exhibits a much more uniform routing distribution. To maintain accuracy, we use a larger average rank budget of 64 and dynamically compensate the top-3 experts, ensuring that the cumulative router score of these leading experts remains sufficiently large to preserve prediction quality.

As shown in Figure \ref{fig:acc_results}, our evaluation across three representative MoE models highlights two observations that validate the design behind our method.

\textit{(1) Uniform low-bit quantization causes accuracy degradation across all MoE architectures.} As shown in our Figure \ref{fig:acc_results}, GPTQ INT2 reduces the average accuracy from 70.03\% to 34.41\% on Mixtral-8×7B, 72.62\% to 37.10\% on Mixtral-8×22B, and 59.63\% to 38.50\% on DeepSeek-MoE-16B. The sharp degradation shows that uniform, aggressive quantization overlooks expert-level heterogeneity and thus conflicts with the dynamic expert usage patterns in MoE models.

\textit{(2) Restoring precision for only a subset of experts is sufficient to recover high accuracy.} In contrast, after applying our method, selectively restoring precision only for the Top-$n$ expert per token yields substantial accuracy recovery under INT2. Specifically, for Mixtral models, selectively restoring only the Top-1 expert brings clear improvements over HQQ on both MMLU and the overall average accuracy. On Mixtral-8×7B, our method increases the INT3 average accuracy by 0.45\% and the INT2 average accuracy by 8.24\%, while the corresponding MMLU improvements are 1.11\% under INT3 and 10.50\% under INT2. Mixtral-8×22B shows a similar pattern: our method improves the INT3 average accuracy by 0.74\% and the INT2 average accuracy by 8.70\%, with MMLU increasing by 3.41\% and 2.83\% respectively. These results indicate that Mixtral’s router distribution is highly skewed, allowing precision restoration for only the dominant expert to recover most of the lost accuracy.

In comparison, DeepSeek-MoE-16B shows noticeably smaller gains over HQQ. our method improves the INT3 average accuracy by 1.63\% and the INT2 average accuracy by 2.30\%, and increases MMLU by 1.83\% under INT3 and 1.73\% under INT2. DeepSeek’s router scores are much more evenly distributed across experts, so restoring only the Top-3 experts yields only marginal benefits. This pattern suggests that DeepSeek requires a higher rank compensator or more experts.

\subsection{Implications on System Performance}
\label{sec:exp:efficiency}

We analyze two representative MoE offloading scenarios and show that our method is compatible and more efficient with existing systems.

\textbf{Case study 1: GPU-only offloading system.}
In this setting, experts and compensators are stored in host memory and loaded into the GPU on demand. Our method integrates seamlessly by transferring only the low-bit experts and compensators of the selected top-$n$ experts, reducing PCIe traffic.

Under \{256,512\} input/output length setting in Figure \ref{fig:thr_results}, Mixtral-Offloading \cite{eliseev2023fast} is constrained by PCIe bandwidth and reaches only 2.37 tokens/s on Mixtral-8$\times$7B and 0.79 tokens/s on Mixtral-8$\times$22B. After integrating our method, throughput increases to 12.27 tokens/s with 3-bit experts and to 18.11 tokens/s with 2-bit experts on Mixtral-8$\times$7B, corresponding to 5.17$\times$ and 7.64$\times$ improvements. Similarly, on Mixtral-8$\times$22B, our method achieves 5.18$\times$ with 3-bit version and 7.63$\times$ with 2-bit version. Hobbit \cite{tang2024hobbit} utilizes mix-precision to optimize the offloading system, but still frequently transfers full-precision experts due to limited cache hit rate. Our method raises throughput from 6.75 tokens/s to 25.43 tokens/s and 36.97 tokens/s on Mixtral-8$\times$7B, providing 3.77$\times$ to 5.48$\times$ acceleration. Improvements on DeepSeek-MoE-16B range from 4.38$\times$ to 5.93$\times$ on Mixtral-Offloading and 2.95$\times$ to 3.78$\times$ on Hobbit, which is smaller than that on Mixtral. This indicates that more activated experts per token increases transfer frequency and limits the system throughput.

\textbf{Case study 2: GPU-NDP offloading system.}
In heterogeneous deployments where experts and compensators reside on NDP devices, our method further reduces bandwidth pressure by executing all non-selected experts directly on NDP using low-bit experts, while only the top-$n$ experts fetch their compensators to the GPU for reconstruction. This enables hybrid execution with lower bandwidth demand.

As shown in Figure \ref{fig:thr_results}, MoNDE \cite{kim2024monde} eliminates most CPU--GPU transfers by executing cold experts directly on NDP. On Mixtral-8$\times$7B, performance improves from 11.56 tokens/s to 54.96 tokens/s with 3-bit experts and to 77.33 tokens/s with 2-bit experts, producing 4.75$\times$ to 6.69$\times$ gains. Mixtral-8$\times$22B exhibits similar ratios, increasing from 3.56 tokens/s to 25.75 tokens/s when using 2-bit experts. DeepSeek-MoE-16B also benefits, with throughput improving from 61.75 tokens/s to 194.51 tokens/s and 233.24 tokens/s, corresponding to 3.15$\times$ to 3.78$\times$ improvements.

Overall, across both GPU-only and GPU-NDP systems, our method consistently provides from 3$\times$ to 8$\times$ end-to-end acceleration. These improvements stem from quantizing the majority of expert computations, which significantly reduces data movement and compute overhead. The cost of transferring compensators for a fixed number of top-ranked experts remains negligible, which we further discuss in Section \ref{sec:exp:ablation}.

\subsection{Ablation Study}
\label{sec:exp:ablation}

\begin{figure}[t]
  \centering

  \includegraphics[width=0.48\textwidth]{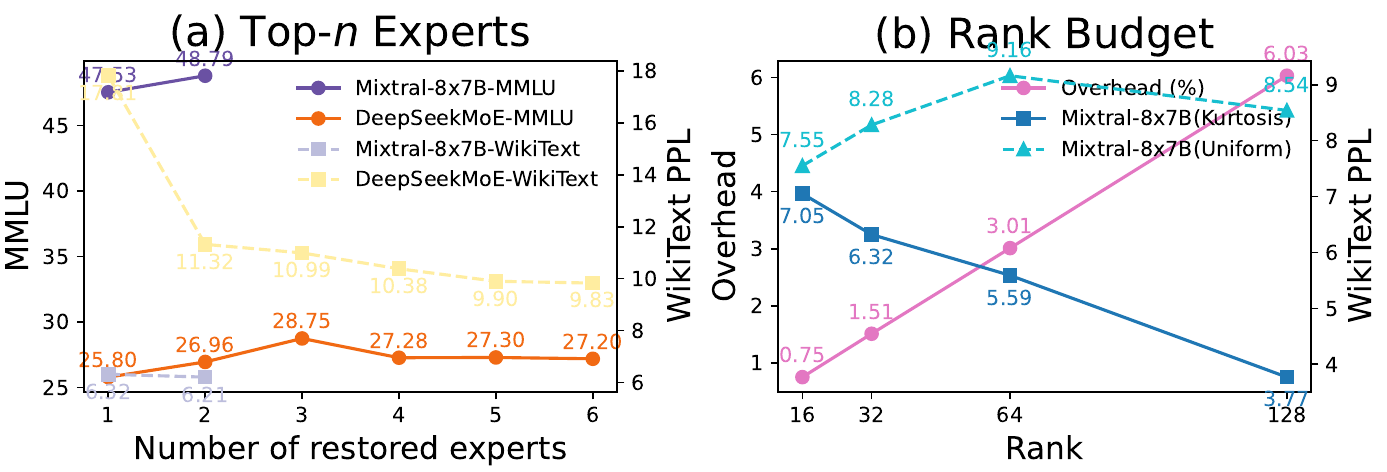}

  \caption{Ablation results on restored expert count, rank budget, and kurtosis-guided allocation.}
  \label{fig:ablation}
\end{figure}

We investigate how our method’s key configs influence model accuracy and system efficiency. Our ablation study follows the 2-bit quantization of our method. Our ablations cover three aspects:

\textit{(1) Number and Position of Restored Experts.} We first study how the number of restored experts affects model quality. As shown in Figure \ref{fig:ablation}(a), for Mixtral-8$\times$7B, restoring only the top-1 expert already recovers most accuracy, and adding the second expert yields only marginal improvement (MMLU from 47.53\% to 48.79\%; WikiText from 6.32 to 6.21). In contrast, DeepSeek-MoE-16B benefits more from increasing the number of restored experts: MMLU increases from 25.80\% (top-1) to 28.75\% (top-3), and WikiText improves from 17.81 to 10.99, but gains saturate beyond three experts.

\begin{table}[h]
\centering
\small
\caption{Model quality when restoring specific experts. Restoring the top-ranked experts yields significantly better accuracy.}
\label{tab:ablation}
\begin{tabular}{|l|c|c|}
\hline
\textbf{Mixtral-8$\times$7B} & \textbf{Only Top-1} & \textbf{Only Top-2} \\ \hline
MMLU (\%)        & 47.53 & 25.26 \\ \hline
WikiText PPL     & 6.32  & 16.76 \\ \hline
\textbf{DeepSeek-MoE-16B} & \textbf{Top1--3} & \textbf{Top-4--6} \\ \hline
MMLU (\%)        & 28.75 & 25.80 \\ \hline
WikiText PPL     & 10.99 & 16.03 \\ \hline
\end{tabular}
\end{table}

To isolate the effect of \emph{position}, we also restore specific expert groups.  
As shown in Table \ref{tab:ablation} For Mixtral-8$\times$7B, restoring only the top-1 expert far outperforms restoring only the top-2 expert (MMLU is 47.53\% and \ 25.26\%; WikiText is 6.32 and 16.76), showing that the top expert dominates contribution. For DeepSeek-MoE-16B, restoring only the top-3 expert outperforms restoring only lower-ranked experts (MMLU: 28.75\% vs.\ 25.80\%; WikiText: 10.99 vs.\ 16.03), indicating that higher-ranked experts are more important.

Overall, these findings validate our motivation: because the router produces skewed importance scores, \textbf{restoring precision for only the high scores experts is essential}.

\textit{(2) Rank Budget on Quality and Overhead.}
We evaluate how the compensator rank influences both quality and bandwidth overhead.
As shown in Figure \ref{fig:ablation}(b), for Mixtral-8×7B, WikiText perplexity decreases as the rank increases from 16 to 128 (PPL from 7.05 to 3.77), showing that higher-rank compensators more effectively correct the quantization residual. This improvement, however, is accompanied by a rise in transfer overhead, growing from 0.32 MB (0.75\% of one INT2 expert) at rank-16 to 2.53 MB (6.03\% of one INT2 expert) at rank-128 per expert.

These results highlight a clear quality–cost trade-off. Increasing the rank beyond 32 yields relatively small additional benefit compared to the extra transfer cost. This findings support our design choice for \textbf{rank selection and efficient compensation}.

\textit{(3) Kurtosis-Guided Rank Allocation vs.\ Uniform Assignment.} We further compare our kurtosis-guided rank allocation with a uniform assignment that gives every expert the same rank in Figure \ref{fig:ablation}(b). Across all rank budgets, the kurtosis-guided strategy consistently achieves lower perplexity on WikiText. At rank-16, it improves from 7.55 for uniform to 7.05. Even at larger rank-128, where both schemes use relatively large compensators, the gap remains substantial (8.54 and 3.77).

These results indicate that experts require different amounts of compensation, and allocating rank based on kurtosis is substantially more effective than uniform assignment. This confirms \textbf{the effectiveness of our kurtosis-guided rank allocation strategy}.

\section{Related Work}

\textbf{Bandwidth-Efficient MoE Inference.} To alleviate GPU memory pressure, recent MoE systems \cite{kamahori2024fiddler,eliseev2023fast} store non-expert weights on GPUs while offloading expert parameters to CPU memory or SSD. However, because different tokens activate different experts, this design triggers frequent on-demand loading and makes MoE inference \textit{memory-bound}, with latency dominated by expert fetching \cite{tang2024hobbit,cao2025moe,zhong2024adapmoe}. Prefetching-based systems \cite{hwang2024pre, song2024promoe, zhong2024adapmoe, cai2024textit} mitigate this bottleneck by fetching experts one layer ahead, but they either rely on auxiliary prediction modules or suffer large performance penalties when predictions are inaccurate. Other approaches reduce data movement through compression, such as sparsity \cite{zhou2025floe} or quantization \cite{tang2024hobbit, duanmu2025mxmoe}. However, these methods still treat experts uniformly and do not address expert-level heterogeneity in routing behavior. GPU–NDP architectures \cite{park2024attacc,kim2024monde,wu2025pimoe, yun2024duplex} further push bandwidth efficiency by enabling in-memory execution of experts, demonstrating a promising commercial trend for MoE acceleration. Our method is compatible with such systems and can complement with further reduction of expert-transfer overhead.

\textbf{Quantization for MoE.} Post-training quantization (PTQ) \cite{dettmers2022gpt3, xiao2023smoothquant, shao2023omniquant, lin2024awq} compresses LLMs without additional training. 
GPTQ \cite{frantar2022gptq} minimizes quantization error via Hessian-guided calibration, while HQQ \cite{badri2023hqq} solves a half-quadratic optimization problem to achieve calibration-free quantization. 
Recent work further adopts a \textit{quantize-then-compensate} paradigm \cite{yao2024exploring,huang2025milo, saha2024compressing, zhang2024lqer}, using low-rank corrections to improve sub-4-bit PTQ. However, when applied to MoE models, existing PTQ methods uniformly quantize all experts and overlook their heterogeneity in routing scores, thus incurring redundancy. This static treatment conflicts with the dynamic nature of MoE inference, making aggressive sub-4-bit quantization unstable. 
Although Hobbit \cite{tang2024hobbit} integrates quantization into MoE system design, its granularity is coarse and does not capture expert-level behavior. Our method offers a system-oriented variant of quantize-then-compensate, selectively restoring a small subset of experts dynamically based on routing score to improve bandwidth efficiency. It complements existing PTQ methods by addressing system-level constraints unique to MoE inference.
\section{Conclusion}

In this work, we implement the quantize–then–compensate paradigm through the lens of bandwidth efficiency and propose a router-guided, selective precision restoration mechanism for MoE inference. Our findings highlight that compensating only a few router-guided experts provides strong accuracy recovery at minimal system cost. Looking forward, future work may explore more model- and hardware-aware rank allocation, improve SVD-based compensators, and develop adaptive strategies for selecting top-$n$ experts, further enhancing both accuracy and system efficiency.

\bibliographystyle{ACM-Reference-Format}
\bibliography{dac_conference}

@article{jiang2024mixtral,
  title={Mixtral of experts},
  author={Jiang, Albert Q and Sablayrolles, Alexandre and Roux, Antoine and Mensch, Arthur and Savary, Blanche and Bamford, Chris and Chaplot, Devendra Singh and Casas, Diego de las and Hanna, Emma Bou and Bressand, Florian and others},
  journal={arXiv preprint arXiv:2401.04088},
  year={2024}
}

@article{liu2024deepseek,
  title={Deepseek-v3 technical report},
  author={Liu, Aixin and Feng, Bei and Xue, Bing and Wang, Bingxuan and Wu, Bochao and Lu, Chengda and Zhao, Chenggang and Deng, Chengqi and Zhang, Chenyu and Ruan, Chong and others},
  journal={arXiv preprint arXiv:2412.19437},
  year={2024}
}

@article{eliseev2023fast,
  title={Fast inference of mixture-of-experts language models with offloading},
  author={Eliseev, Artyom and Mazur, Denis},
  journal={arXiv preprint arXiv:2312.17238},
  year={2023}
}

@article{tang2024hobbit,
  title={Hobbit: A mixed precision expert offloading system for fast moe inference},
  author={Tang, Peng and Liu, Jiacheng and Hou, Xiaofeng and Pu, Yifei and Wang, Jing and Heng, Pheng-Ann and Li, Chao and Guo, Minyi},
  journal={arXiv preprint arXiv:2411.01433},
  year={2024}
}

@inproceedings{cao2025moe,
  title={Moe-lightning: High-throughput moe inference on memory-constrained gpus},
  author={Cao, Shiyi and Liu, Shu and Griggs, Tyler and Schafhalter, Peter and Liu, Xiaoxuan and Sheng, Ying and Gonzalez, Joseph E and Zaharia, Matei and Stoica, Ion},
  booktitle={Proceedings of the 30th ACM International Conference on Architectural Support for Programming Languages and Operating Systems, Volume 1},
  pages={715--730},
  year={2025}
}

@inproceedings{hwang2024pre,
  title={Pre-gated moe: An algorithm-system co-design for fast and scalable mixture-of-expert inference},
  author={Hwang, Ranggi and Wei, Jianyu and Cao, Shijie and Hwang, Changho and Tang, Xiaohu and Cao, Ting and Yang, Mao},
  booktitle={2024 ACM/IEEE 51st Annual International Symposium on Computer Architecture (ISCA)},
  pages={1018--1031},
  year={2024},
  organization={IEEE}
}

@inproceedings{zhong2024adapmoe,
  title={AdapMoE: Adaptive sensitivity-based expert gating and management for efficient moe inference},
  author={Zhong, Shuzhang and Liang, Ling and Wang, Yuan and Wang, Runsheng and Huang, Ru and Li, Meng},
  booktitle={Proceedings of the 43rd IEEE/ACM International Conference on Computer-Aided Design},
  pages={1--9},
  year={2024}
}

@article{song2024promoe,
  title={Promoe: Fast moe-based llm serving using proactive caching},
  author={Song, Xiaoniu and Zhong, Zihang and Chen, Rong and Chen, Haibo},
  journal={arXiv preprint arXiv:2410.22134},
  year={2024}
}

@article{cai2024textit,
  title={Read-ME: Refactorizing LLMs as Router-Decoupled Mixture of Experts with System Co-Design},
  author={Cai, Ruisi and Ro, Yeonju and Kim, Geon-Woo and Wang, Peihao and Ehteshami Bejnordi, Babak and Akella, Aditya and Wang, Zhangyang and others},
  journal={Advances in Neural Information Processing Systems},
  volume={37},
  pages={116126--116148},
  year={2024}
}

@article{frantar2022gptq,
  title={Gptq: Accurate post-training quantization for generative pre-trained transformers},
  author={Frantar, Elias and Ashkboos, Saleh and Hoefler, Torsten and Alistarh, Dan},
  journal={arXiv preprint arXiv:2210.17323},
  year={2022}
}

@misc{badri2023hqq,
title  = {Half-Quadratic Quantization of Large Machine Learning Models},
url    = {https://mobiusml.github.io/hqq_blog/},
author = {Hicham Badri and Appu Shaji},
month  = {November},
year   = {2023}
}

@article{dettmers2022gpt3,
  title={Gpt3. int8 (): 8-bit matrix multiplication for transformers at scale},
  author={Dettmers, Tim and Lewis, Mike and Belkada, Younes and Zettlemoyer, Luke},
  journal={Advances in neural information processing systems},
  volume={35},
  pages={30318--30332},
  year={2022}
}

@inproceedings{xiao2023smoothquant,
  title={Smoothquant: Accurate and efficient post-training quantization for large language models},
  author={Xiao, Guangxuan and Lin, Ji and Seznec, Mickael and Wu, Hao and Demouth, Julien and Han, Song},
  booktitle={International conference on machine learning},
  pages={38087--38099},
  year={2023},
  organization={PMLR}
}

@article{shao2023omniquant,
  title={Omniquant: Omnidirectionally calibrated quantization for large language models},
  author={Shao, Wenqi and Chen, Mengzhao and Zhang, Zhaoyang and Xu, Peng and Zhao, Lirui and Li, Zhiqian and Zhang, Kaipeng and Gao, Peng and Qiao, Yu and Luo, Ping},
  journal={arXiv preprint arXiv:2308.13137},
  year={2023}
}

@article{lin2024awq,
  title={Awq: Activation-aware weight quantization for on-device llm compression and acceleration},
  author={Lin, Ji and Tang, Jiaming and Tang, Haotian and Yang, Shang and Chen, Wei-Ming and Wang, Wei-Chen and Xiao, Guangxuan and Dang, Xingyu and Gan, Chuang and Han, Song},
  journal={Proceedings of machine learning and systems},
  volume={6},
  pages={87--100},
  year={2024}
}

@inproceedings{yao2024exploring,
  title={Exploring post-training quantization in llms from comprehensive study to low rank compensation},
  author={Yao, Zhewei and Wu, Xiaoxia and Li, Cheng and Youn, Stephen and He, Yuxiong},
  booktitle={Proceedings of the AAAI Conference on Artificial Intelligence},
  volume={38},
  number={17},
  pages={19377--19385},
  year={2024}
}

@inproceedings{kim2024monde,
  title={Monde: Mixture of near-data experts for large-scale sparse models},
  author={Kim, Taehyun and Choi, Kwanseok and Cho, Youngmock and Cho, Jaehoon and Lee, Hyuk-Jae and Sim, Jaewoong},
  booktitle={Proceedings of the 61st ACM/IEEE Design Automation Conference},
  pages={1--6},
  year={2024}
}

@misc{eval-harness,
  author       = {Gao, Leo and Tow, Jonathan and Abbasi, Baber and Biderman, Stella and Black, Sid and DiPofi, Anthony and Foster, Charles and Golding, Laurence and Hsu, Jeffrey and Le Noac'h, Alain and Li, Haonan and McDonell, Kyle and Muennighoff, Niklas and Ociepa, Chris and Phang, Jason and Reynolds, Laria and Schoelkopf, Hailey and Skowron, Aviya and Sutawika, Lintang and Tang, Eric and Thite, Anish and Wang, Ben and Wang, Kevin and Zou, Andy},
  title        = {The Language Model Evaluation Harness},
  month        = 07,
  year         = 2024,
  publisher    = {Zenodo},
  version      = {v0.4.3},
  doi          = {10.5281/zenodo.12608602},
  url          = {https://zenodo.org/records/12608602}
}

@inproceedings{amini2019mathqa,
  title={Mathqa: Towards interpretable math word problem solving with operation-based formalisms},
  author={Amini, Aida and Gabriel, Saadia and Lin, Shanchuan and Koncel-Kedziorski, Rik and Choi, Yejin and Hajishirzi, Hannaneh},
  booktitle={Proceedings of the 2019 conference of the North American chapter of the association for computational linguistics: Human language technologies, volume 1 (long and short papers)},
  pages={2357--2367},
  year={2019}
}

@article{zellers2019hellaswag,
  title={Hellaswag: Can a machine really finish your sentence?},
  author={Zellers, Rowan and Holtzman, Ari and Bisk, Yonatan and Farhadi, Ali and Choi, Yejin},
  journal={arXiv preprint arXiv:1905.07830},
  year={2019}
}

@article{clark2018think,
  title={Think you have solved question answering? try arc, the ai2 reasoning challenge},
  author={Clark, Peter and Cowhey, Isaac and Etzioni, Oren and Khot, Tushar and Sabharwal, Ashish and Schoenick, Carissa and Tafjord, Oyvind},
  journal={arXiv preprint arXiv:1803.05457},
  year={2018}
}

@article{clark2019boolq,
  title={Boolq: Exploring the surprising difficulty of natural yes/no questions},
  author={Clark, Christopher and Lee, Kenton and Chang, Ming-Wei and Kwiatkowski, Tom and Collins, Michael and Toutanova, Kristina},
  journal={arXiv preprint arXiv:1905.10044},
  year={2019}
}

@inproceedings{bisk2020piqa,
  title={Piqa: Reasoning about physical commonsense in natural language},
  author={Bisk, Yonatan and Zellers, Rowan and Gao, Jianfeng and Choi, Yejin and others},
  booktitle={Proceedings of the AAAI conference on artificial intelligence},
  volume={34},
  number={05},
  pages={7432--7439},
  year={2020}
}

@article{sakaguchi2021winogrande,
  title={Winogrande: An adversarial winograd schema challenge at scale},
  author={Sakaguchi, Keisuke and Bras, Ronan Le and Bhagavatula, Chandra and Choi, Yejin},
  journal={Communications of the ACM},
  volume={64},
  number={9},
  pages={99--106},
  year={2021},
  publisher={ACM New York, NY, USA}
}

@article{huang2025milo,
  title={MiLo: Efficient Quantized MoE Inference with Mixture of Low-Rank Compensators},
  author={Huang, Beichen and Yuan, Yueming and Shao, Zelei and Zhang, Minjia},
  journal={arXiv preprint arXiv:2504.02658},
  year={2025}
}

@article{chmiel2020robust,
  title={Robust quantization: One model to rule them all},
  author={Chmiel, Brian and Banner, Ron and Shomron, Gil and Nahshan, Yury and Bronstein, Alex and Weiser, Uri and others},
  journal={Advances in neural information processing systems},
  volume={33},
  pages={5308--5317},
  year={2020}
}

@article{akhondzadeh2025kurtail,
  title={KurTail: Kurtosis-based LLM Quantization},
  author={Akhondzadeh, Mohammad Sadegh and Bojchevski, Aleksandar and Eleftheriou, Evangelos and Dazzi, Martino},
  journal={arXiv preprint arXiv:2503.01483},
  year={2025}
}

@article{huang2024mixture,
  title={Mixture Compressor for Mixture-of-Experts LLMs Gains More},
  author={Huang, Wei and Liao, Yue and Liu, Jianhui and He, Ruifei and Tan, Haoru and Zhang, Shiming and Li, Hongsheng and Liu, Si and Qi, Xiaojuan},
  journal={arXiv preprint arXiv:2410.06270},
  year={2024}
}

@inproceedings{chitty2025mopeq,
  title={MoPEQ: Mixture of Mixed Precision Quantized Experts},
  author={Chitty-Venkata, Krishna Teja and Ye, Jie and Emani, Murali},
  booktitle={Proceedings of the IEEE/CVF International Conference on Computer Vision},
  pages={4023--4032},
  year={2025}
}

@inproceedings{wu2025pimoe,
  title={PIMoE: Towards efficient MoE transformer deployment on NPU-PIM system through throttle-aware task offloading},
  author={Wu, Lizhou and Zhu, Haozhe and He, Siqi and Lin, Xuanda and Zeng, Xiaoyang and Chen, Chixiao},
  booktitle={2025 62nd ACM/IEEE Design Automation Conference (DAC)},
  pages={1--7},
  year={2025},
  organization={IEEE}
}

@article{kamahori2024fiddler,
  title={Fiddler: Cpu-gpu orchestration for fast inference of mixture-of-experts models},
  author={Kamahori, Keisuke and Tang, Tian and Gu, Yile and Zhu, Kan and Kasikci, Baris},
  journal={arXiv preprint arXiv:2402.07033},
  year={2024}
}

@article{zhou2025floe,
  title={FloE: On-the-Fly MoE Inference on Memory-constrained GPU},
  author={Zhou, Yuxin and Li, Zheng and Zhang, Jun and Wang, Jue and Wang, Yiping and Xie, Zhongle and Chen, Ke and Shou, Lidan},
  journal={arXiv preprint arXiv:2505.05950},
  year={2025}
}

@article{duanmu2025mxmoe,
  title={MxMoE: Mixed-precision Quantization for MoE with Accuracy and Performance Co-Design},
  author={Duanmu, Haojie and Li, Xiuhong and Yuan, Zhihang and Zheng, Size and Duan, Jiangfei and Zhang, Xingcheng and Lin, Dahua},
  journal={arXiv preprint arXiv:2505.05799},
  year={2025}
}

@article{yang2025qwen3,
  title={Qwen3 technical report},
  author={Yang, An and Li, Anfeng and Yang, Baosong and Zhang, Beichen and Hui, Binyuan and Zheng, Bo and Yu, Bowen and Gao, Chang and Huang, Chengen and Lv, Chenxu and others},
  journal={arXiv preprint arXiv:2505.09388},
  year={2025}
}

@article{shazeer2017outrageously,
  title={Outrageously large neural networks: The sparsely-gated mixture-of-experts layer},
  author={Shazeer, Noam and Mirhoseini, Azalia and Maziarz, Krzysztof and Davis, Andy and Le, Quoc and Hinton, Geoffrey and Dean, Jeff},
  journal={arXiv preprint arXiv:1701.06538},
  year={2017}
}

@article{muennighoff2024olmoe,
  title={Olmoe: Open mixture-of-experts language models},
  author={Muennighoff, Niklas and Soldaini, Luca and Groeneveld, Dirk and Lo, Kyle and Morrison, Jacob and Min, Sewon and Shi, Weijia and Walsh, Pete and Tafjord, Oyvind and Lambert, Nathan and others},
  journal={arXiv preprint arXiv:2409.02060},
  year={2024}
}

@inproceedings{park2024attacc,
  title={Attacc! unleashing the power of pim for batched transformer-based generative model inference},
  author={Park, Jaehyun and Choi, Jaewan and Kyung, Kwanhee and Kim, Michael Jaemin and Kwon, Yongsuk and Kim, Nam Sung and Ahn, Jung Ho},
  booktitle={Proceedings of the 29th ACM International Conference on Architectural Support for Programming Languages and Operating Systems, Volume 2},
  pages={103--119},
  year={2024}
}

@inproceedings{yun2024duplex,
  title={Duplex: A device for large language models with mixture of experts, grouped query attention, and continuous batching},
  author={Yun, Sungmin and Kyung, Kwanhee and Cho, Juhwan and Choi, Jaewan and Kim, Jongmin and Kim, Byeongho and Lee, Sukhan and Sohn, Kyomin and Ahn, Jung Ho},
  booktitle={2024 57th IEEE/ACM International Symposium on Microarchitecture (MICRO)},
  pages={1429--1443},
  year={2024},
  organization={IEEE}
}

@article{hendrycks2020measuring,
  title={Measuring massive multitask language understanding},
  author={Hendrycks, Dan and Burns, Collin and Basart, Steven and Zou, Andy and Mazeika, Mantas and Song, Dawn and Steinhardt, Jacob},
  journal={arXiv preprint arXiv:2009.03300},
  year={2020}
}

@article{raffel2020exploring,
  title={Exploring the limits of transfer learning with a unified text-to-text transformer},
  author={Raffel, Colin and Shazeer, Noam and Roberts, Adam and Lee, Katherine and Narang, Sharan and Matena, Michael and Zhou, Yanqi and Li, Wei and Liu, Peter J},
  journal={Journal of machine learning research},
  volume={21},
  number={140},
  pages={1--67},
  year={2020}
}

@article{kim2015ramulator,
  title={Ramulator: A fast and extensible DRAM simulator},
  author={Kim, Yoongu and Yang, Weikun and Mutlu, Onur},
  journal={IEEE Computer architecture letters},
  volume={15},
  number={1},
  pages={45--49},
  year={2015},
  publisher={IEEE}
}

@article{saha2024compressing,
  title={Compressing large language models using low rank and low precision decomposition},
  author={Saha, Rajarshi and Sagan, Naomi and Srivastava, Varun and Goldsmith, Andrea and Pilanci, Mert},
  journal={Advances in Neural Information Processing Systems},
  volume={37},
  pages={88981--89018},
  year={2024}
}

@article{zhang2024lqer,
  title={Lqer: Low-rank quantization error reconstruction for llms},
  author={Zhang, Cheng and Cheng, Jianyi and Constantinides, George A and Zhao, Yiren},
  journal={arXiv preprint arXiv:2402.02446},
  year={2024}
}

\end{document}